\pdfoutput=1

\documentclass[11pt]{article}

\usepackage[final]{acl}

\usepackage{times}
\usepackage{latexsym}

\usepackage[T1]{fontenc}

\usepackage[utf8]{inputenc}

\usepackage{microtype}

\usepackage{inconsolata}

\usepackage{graphicx}
\usepackage{amsmath}
\usepackage{xcolor}
\usepackage{booktabs}
\usepackage{multirow}
\usepackage{enumitem}

\usepackage{dblfloatfix}
\usepackage{multicol}
\usepackage{tikz}
\usepackage{pgfplots}

\newcommand{\round}[1]{\ensuremath{\lfloor#1\rceil}}

\newcommand{\naming}{GIFT-SW{}}

\title{GIFT-SW: Gaussian noise Injected Fine-Tuning of Salient Weights for LLMs}

\author{
 \textbf{Maxim Zhelnin$^\clubsuit$ \textsuperscript{1}}, 
 \textbf{Viktor Moskvoretskii$^\clubsuit$ \textsuperscript{1,3}},
 \textbf{Egor Shvetsov\textsuperscript{1}}, \\
 \textbf{Egor Venediktov},
 \textbf{Mariya Krylova},
  \textbf{Aleksandr Zuev},
 \textbf{Evgeny Burnaev \textsuperscript{1,2}}
\\
 \textsuperscript{1} \small{Skolkovo Institute of Science and Technology} \\
 \textsuperscript{2} \small{Artificial Intelligence Research Institute} \\
 \textsuperscript{3} \small{HSE University}
\\
 \small{
   \textbf{Correspondence:} \href{mailto: m.zhelnin@skol.tech}{m.zhelnin@skol.tech}
 }
 \small{ $\clubsuit$ indicates equal contribution.}
}
\begin{document}

\maketitle
\begin{abstract}

Parameter Efficient Fine-Tuning (PEFT) methods have gained popularity and democratized the usage of Large Language Models (LLMs). 
Recent studies have shown that a small subset of weights significantly impacts performance.
Based on this observation, we introduce a novel PEFT method, called Gaussian noise Injected Fine Tuning of Salient Weights (\naming{}). 
Our method updates only salient columns, while injecting Gaussian noise into non-salient ones.
To identify these columns, we developed a generalized sensitivity metric that extends and unifies metrics from previous studies. Experiments with LLaMA models demonstrate that \naming{} outperforms full fine-tuning and modern PEFT methods under the same computational budget. Moreover, GIFT-SW offers practical advantages to recover performance of models subjected to mixed-precision quantization with keeping salient weights in full precision. Code is available in \href{https://github.com/On-Point-RND/GIFT_SW}{our repository}.

\end{abstract}

\section{Introduction}
\hspace{-1em}
\begin{figure*}[t!]
    \centering
    \includegraphics[width=\textwidth]{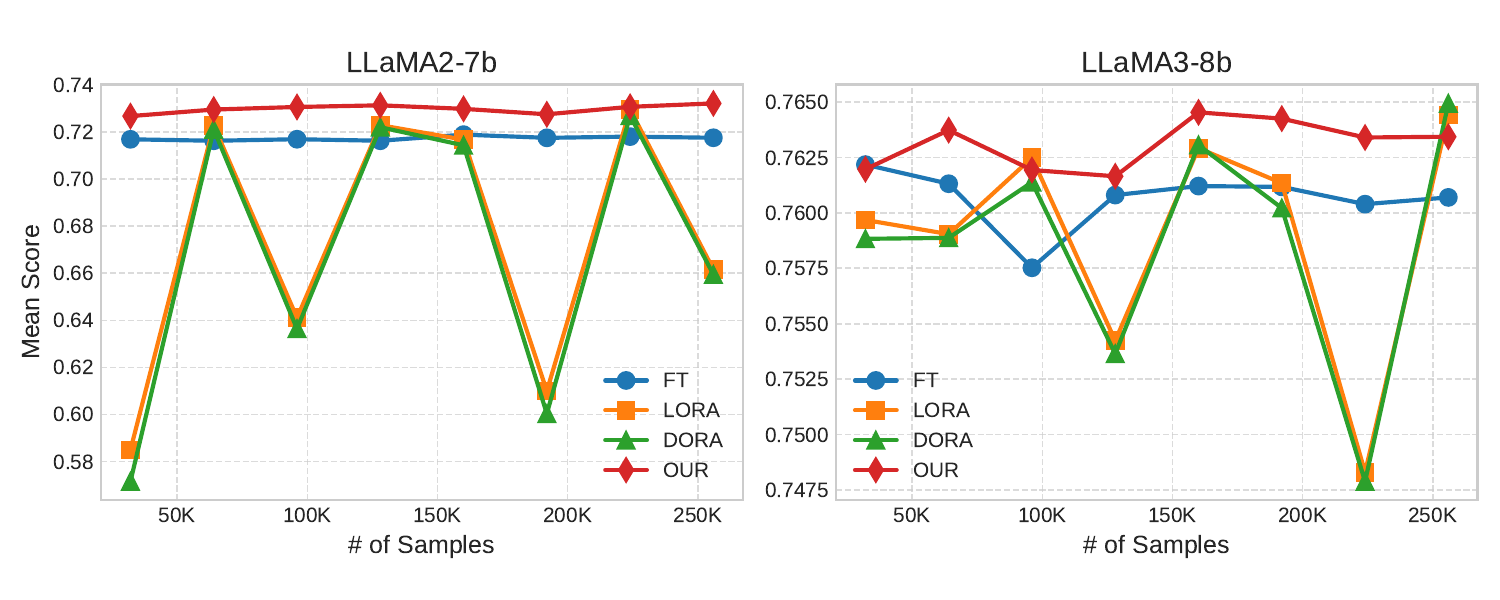}
    \caption{Mean performance of different fine-tuning approaches for LLaMA models with scaling data budget. \naming{} shows superior performance with nearly all data budgets, also being as stable as full fine-tuning.}
    \label{fig:performance_comparison}
\end{figure*}

Modern LLMs demonstrate remarkable generalization capabilities on unseen tasks. However, fine-tuning remains crucial to enhance these models performance or to restore the performance after compression techniques like quantization~\cite{dettmers2024qlora,moskvoretskii-etal-2024-large}, pruning~\cite{frantar2023sparsegpt, kim2023squeezellm}, or tensor decomposition have been applied. Given the large scale of modern LLMs, fine-tuning all parameters can be computationally and memory-intensive. To overcome this challenge, Parameter Efficient Fine-Tuning schemes have been developed, aimed to improve model performance while using limited computational and memory resources.

To date, PEFT methods have not matched the accuracy of full fine-tuning~\cite{nikdan2024rosa}, highlighting the need for new approaches that can close this gap while still minimizing resource use. Additionally, most PEFT methods involve adding extra parameters, which increases computational demands.

\label{sec:questions}
To address those issues and enhance the performance of efficiently trained LLMs, we introduce a novel PEFT method, \naming{}. This approach focuses on updating a small subset of salient weights while injecting noise into the non-salient weights. The development of this method is grounded in observations from previous studies and the related questions they raise, which we aim to answer:

Previous research has shown that there is a small subset of salient weights which can significantly affect the effectiveness of post-training quantization (PTQ) \cite{dettmers2022gpt3, dettmers2023spqr, kim2023squeezellm} and pruning techniques \cite{yin2023outlier, frantar2023sparsegpt, sun2023simple}. Moreover, \citeauthor{gurnee2024universal} identified a group of "universal neurons" that are critical to a model's functionality, emphasizing the importance of selecting and updating these salient weights.
\textbf{\textit{Question 1: Does updating a small subset of salient weights is sufficient to adjust the model?}}

Recent studies have demonstrated that Perturbed Gradient Descent (PGD), with noise injections applied both before and after the gradient step, can stabilize convergence and help prevent overfitting \cite{poole2014analyzing, zhu2018anisotropic, jin2021nonconvex}. 
\textbf{\textit{Question 2: Does Injecting Noise helps convergence?}}

PGD is commonly employed to enhance model robustness by approximating the quantization process \cite{shvetsov2022quantnas, shin2023nipq, defossez2021differentiable}. This increased robustness can aid in maintaining the quality of the quantized model.
\textbf{\textit{Question 3: Does injecting noise helps robustness?}}






Selecting salient weights is a significant challenge, particularly in quantization and pruning, and it is central to our method. In our paper, we derive a general formulation for all previously established saliency metrics and present experiments to compare their effectiveness.



The main contributions of our work can be summarized as follows:
\begin{itemize}
    \item We introduce a novel PEFT method for pre-trained and quantized LLMs, called \naming{}. It is designed to fine-tune weights in salient columns while injecting Gaussian noise into non-salient weights, which are kept frozen during training.
    \item We generalize sensitivity metrics for identifying salient columns in pre-trained LLMs. We compare various novel and existing instances of the proposed general form and identify a new metric,  which on average outperform  previously studied in the literature metrics\cite{xiao2023smoothquant,lee2024owq}. 
    \item Experiments demonstrate that \naming{} outperforms modern PEFT methods and full fine-tuning baselines across most zero-shot tasks. \naming{} for LLaMA models achieve comparable accuracy to the corresponding state-of-the-art T\"ULU2 models, despite fine-tuning only 3\% of the parameters and utilizing ten times less computational resources.
    \item We demonstrate that \naming{} is more stable with respect to a size of training set compared with low-rank adapters.
\end{itemize}




\section{Related Work}
\subsection{Parameter efficient fine-tuning of LLM}
One of the most popular method with high efficiency is LoRA \cite{hu2021lora}, which trains the low-rank adapters. 
Recent modifications to the method aim to improve the initialization of the adapters \cite{liu2024dora} and enhance the low-rank representation of pre-trained weights by adding sparse adapters \cite{nikdan2024rosa}. Another improvement of the learning capacity of LoRA is given by DoRA \cite{liu2024dora}, which fine-tunes magnitude and direction components of the pretrained weights. This method achieves considerable performance across various fine-tuning tasks.




\subsection{Salient Weights in LLMs}
The identification of salient weights\protect\footnote{In our work, we use the terms \textbf{salient weights} and weight \textbf{outliers} interchangeably.} is one of the main problems in weight pruning. Recently, several approaches have been proposed to identify such weights in LLMs, including SparseGPT~\cite{frantar2023sparsegpt}, Wanda~\cite{sun2023simple}, and OWL~\cite{yin2023outlier}.

\citeposs{dettmers2022gpt3} demonstrated that a small subset of outliers in input activations has a substantial impact on LLM performance, highlighting the relationship between the activation outliers and the salient weights. Many subsequent Post-Training Quantization (PTQ) methods used similar or identical pruning metrics to identify these salient weights~\cite{dettmers2023spqr, xiao2023smoothquant, lee2024owq}. 

In our work, we generalize the identification metrics for salient weights by considering metrics from both the literature on pruning and quantization.

\subsection{Structured and Non-structured Salient Weights selection}
Since salient weights account for only a few percent of all the weights, a straightforward approach to preserve them would be to store unstructured salient weights in a sparse matrix. \cite{dettmers2023spqr} demonstrated that this approach is computationally reasonable and leads to performance improvement. On the other hand, \citeposs{xiao2023smoothquant} revealed that outliers in activations are confined to a small fraction of weight channels, which was incorporated into SmoothQuant, where outlier columns are identified using a small calibration dataset. This concept is further developed in QUIK~\cite{ashkboos2023quik}, where outlier columns are retained in full precision, while other columns are quantized using GPTQ~\citep{frantar2022gptq}. A similar procedure is used in OWQ~\cite{lee2024owq}, but with an OBD-based metric~\cite{lecun1989optimal}.

Due to the lack of results in the literature on which approach brings better results, structured or unstructured salient weight selection, and motivated by computational efficiency mentioned in ~\cite{ashkboos2023quik}, in our work we follow the second line of work with structured column-wise salient weight selection.

\subsection{Noise Injections}
In this section, we briefly describe Gaussian Noise Injections (GNI) and its benefits. Then, we show that the approximation of quantization noise and GNI are identical. Therefore, GNI can also benefit further model quantization. Therefor, to examine our third question, we sample noise relative to quantization levels, leaving other sampling options for future work.


\textbf{Gaussian Noise Injections (GNI)}. Perturbed Gradient Descent (PGD) is a family of methods that involve adding or multiplying weights with samples from some random distribution, during an optimization procedure. 
Gaussian noise injection (GNI) after the gradient step helps model to escape saddle points efficiently in non-convex optimization~\citep{jin2021nonconvex}. However, when Gaussian noise is injected before the gradient step, it helps model to escape from the spurious local optimum~\citep{zhu2018anisotropic}.


\begin{align}
    \label{eq:pgd_one} \theta_{t+1} \leftarrow \theta_{t} - \tau (\nabla f(\theta_{t}) + \xi) \\\ 
    \label{eq:pgd_two} \theta_{t+1} \leftarrow \theta_{t} - \tau (\nabla f(\theta_{t} + \xi)  ) \\
    \xi \sim \mathcal{N}(\mu,\,\sigma^{2})  
\end{align}

Moreover, practical benefits of noise injections are well documented in the literature and often can be discussed as regularization techniques~\cite{bishop1995training, srivastava2014dropout, camuto2020explicit}, methods to prompt adversarial robustenss~\cite{panda2021implicit} and to be used for data agumentation~\cite{moreno2018forward}.


 
In our work we use GNI \textit{before evaluating the gradient}. For this scenario, 
\citet{orvieto2023explicit} proposed to add noise only to one layer at training iteration to avoid variance explosion. It was empirically and theoretically demonstrated that GNI 
serves as a regularization.   
\citet{liu2023pac} study fine-tuning of pre-trained Language Models with GNI. Authors propose first to learn layer-wise variance parameters for noise distributions and then to fine-tune the model by adding noise to all the weights. The obtained results showed that the approach is superior to independent layer-wise noise injections.




\textbf{Quantization Noise Injections (QNI).}\label{sec:background_QNI} Quantization aware training (QAT) of networks is applied to mitigate their accuracy degradation after quantization. However, uniform quantization~\footnote{For the reader not familiar with uniform quantization, we discuss it in more details in Section~\ref{appendix:uniform quantization}.} $Q$ is a non-differentiable operation. For simplicity, it can be expressed as a composition of scaling and rounding operations, $Q(\mathbf{W})=\Delta\round{\frac{\mathbf{W}}{\Delta}}$. In terms of QAT operation $Q$ can be efficiently approximated with quantization noise $\mathbf{\Omega}$ such that $\mathbf{\Omega} = Q(\mathbf{W}) - \mathbf{W}$~\cite{defossez2021differentiable, shvetsov2022quantnas, shin2023nipq}. Thus, training models with QNI is exactly the same as employing PGD with GNI before evaluating the gradient.

Under some assumptions the noise $\mathbf{\Omega}$ induced by uniform quantization can often be modeled by an additive noise that is uniformly distributed, uncorrelated with the input signal, and has a white spectrum \cite{widrow1996statistical}. However in practice, the conditions are often not satisfied. Therefore employing Gaussian distribution $\mathcal{N}(\mu,\,\sigma^{2})$ for $\mathbf{\Omega}$  typically yields improved outcomes \cite{defossez2021differentiable, shvetsov2022quantnas}. 

Although GNI is beneficial for model training there is no clear answer on how to choose noise parameters. \citet{liu2023pac} determine noise parameters such that KL divergence between original and perturbed weights is minimized.  \citet{shin2023nipq} identify parameters of the Gaussian distribution to resemble the weight distribution with a scale proportional to quantization step.


\subsection{Straight Through Estimator}
The most popular QAT technique incorporating quantization operation into the traning process is Straight Through Estimation (STE)\footnote{More details on STE can be found in Section~\ref{appendix: STE}.}~\cite{bengio2013estimating, shang2023pb_llm}, which basically re-parameterizes gradients. However, \citeposs{defossez2021differentiable} demonstrated that STE has some disadvantages compared with QNI\footnote{Event though QNI and GNI are identical operations for consistency and clarity,  in the case of quantization we will refer to this procedure as Quantization Noise Injections (QNI)}, as STE is biased and may cause weight oscillation between quantization steps.  \citeposs{shin2023nipq} demonstrated that pretraining models for the following quantization with QNI instead of STE results in better performance. More technical details are provided in Section~\ref{appendix: STE}.



\section{Method}

\begin{figure}[t!]
    \centering
    \includegraphics[width=0.5\textwidth]{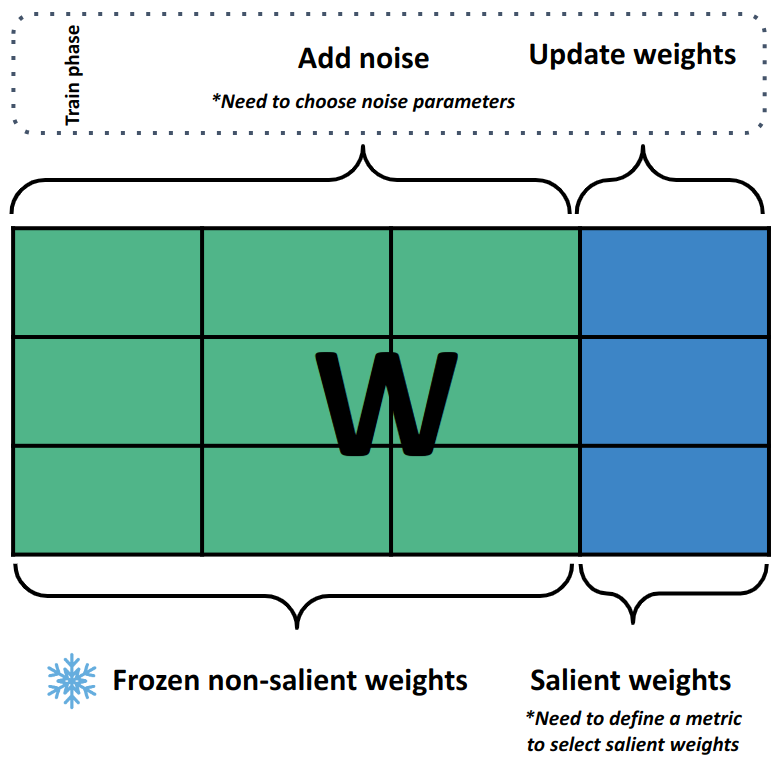}
    \caption{\naming{} procedure follows Equation~\ref{eq:pgd_two}. We first sample some noise, relative to quantization levels, then, perform forward pass, and then update
    salient weights only. In \naming{},  quantization, pruning or tensor decomposition can be applied to non-salient weights and then, salient weights can be fine-tuned effectively without changing non-salient weights structure. In our experiments we select only 128 columns of salient weights, unless specified otherwise.}
    \label{fig:gift_sw_scheme}
\end{figure}

\naming{} consists of the following steps:
 \vspace{-2mm}
\begin{itemize}
\setlength\itemsep{0em}
\item[(1)] Identify a fixed number of salient columns using a chosen sensitive metric, based on a small calibration set. This number remains consistent across all layers.  \vspace{-2mm}
\item[(2)] Split columns of the matrices into subsets of salient columns and regular ones.  
\vspace{-2mm}
\item[(3)] During training, add noise to the weights in non-salient columns and update weights only in the salient columns.  
\vspace{-2mm}
\end{itemize}

Thus, the method depends on two main design choices: 1) how to choose salient columns and 2) the parameters of noise injections. We cover the choice of metrics in Section~\ref{sec:metrics}. Noise injection details are provided in Section~\ref{sec:noise}.

\subsection{Generalizing parameter sensitivity metrics}
\label{sec:metrics}
Several approaches have been proposed recently to identify weights sensitive to quantization~\cite{dettmers2023spqr} or pruning~\cite{sun2023simple}. We generalize them as metrics for sensitivity to perturbations, and by applying these metrics, we determine which columns are more susceptible to degradation. Therefore, we avoid adding noise to such columns and use them to fine-tune the model.



The proposed sensitivity metric is written for a column $j$ of weight matrix $\mathbf{W}$ as
\begin{equation}
  \label{eq:our_metric}
  s_{j} = \Vert \mathbf{D}_j \Vert_{\tau} \Vert \mathbf{X}_{j} \Vert_{\rho}^{\gamma}, 
\end{equation}
where $\mathbf{D}_j$ is a measure of weights perturbation, $s_{j}$ denotes sensitivity of the column to perturbations,  $\mathbf{X}$ is the input feature, and $\gamma$ takes on one of the following values $1/2, 1, 2$.  As discussed in Section~\ref{sec:background_QNI} we could apply GNI as a source of  perturbations, then we would compute  $\mathbf{D}_j = \mathbf{W}_{:, j} +  \xi $. However, sampling  noise $\xi$ is not deterministic. To approximate an influence of the noise $\xi$ we utilize perturbations caused by quantization.\footnote{Optionally, one could use weight pruning as a source of perturbations or any other.}
That would lead to $\mathbf{D}_j= \mathbf{W}_{:, j} - Q(\mathbf{W}_{:, j})$, where  $Q(\mathbf{W}_{:, j})$ 
corresponds to the weights subjected to uniform symmetric quantization (see Appendix \ref{appendix:uniform quantization}).

The input feature $\mathbf{X}$ for each layer is computed using a number of random sentences from a calibration dataset. After that, sensitivity values $s_{j}$ are estimated for individual columns. Columns with the highest values are identified as the salient columns. Some details about the calibration dataset is described in Section \ref{sec:exps_data}.

The metric given by Equation~\ref{eq:our_metric} is closely related to those studied in the recent literature on quantization. In particular, the metric $\Vert \mathbf{X}\Vert_{\infty}$ is employed in QUIK \cite{ashkboos2023quik} and SmoothQuant \cite{xiao2023smoothquant}. OWQ \cite{lee2024owq} adopts $\lambda_{j}\Vert \mathbf{D}_j\Vert_{2}^{2}$, where $\lambda_{j} = \Vert \mathbf{X}_{j} \Vert_{2}^{2}$ is the $j$-th diagonal element of the Hessian matrix $\mathbf{H}$ for the layer quantization error. It can be seen, that the sensitivity metric used in OWQ is a modification for column quantization of the salience measure provided in OBD \cite{lecun1989optimal} for network pruning. A metric proposed in Wanda \cite{sun2023simple} is element-wise variant of the metric $\Vert \mathbf{D}_j \Vert_1 \Vert \mathbf{X}_{j} \Vert_{2} $, which can be easily obtained from Equation~\ref{eq:our_metric} with pruning as a source of perturbations for $\mathbf{D}_j$.



In contrast to Wanda, we use $l_{\infty}$ norm in our general Equation ~\ref{eq:our_metric} due to the following observations, examples contained in a calibration dataset induce different values of the input feature, a use of $l_{2}$ norm leads to averaging of the values along input channels. Therefore, the appearance of the outlier values in the input activation can be obscured by a large number of lower values. The same conclusions can be also applied to the weight error. In the case of the $l_2$ norm, the error for each channel includes all deviations between the quantized and original weights. Therefore, rare considerable errors can be mitigated by a large number of small deviations.

\subsection{Quantization Noise Injection}
\label{sec:noise}
To improve our fine-tuning procedure with QNI, we avoid applying perturbations to sensitive weights. Therefore, after identifying columns that are sensitive to perturbations or salient during the fine-tuning stage, we inject quantization noise only into non-salient columns across all layers, as shown in Figure~\ref{fig:gift_sw_scheme}. 

The scale parameters of the Gaussian noise are determined by the quantization step sizes, which are computed for each layer prior to the training process. 

For the weight matrix $\mathbf{W}$ of a given layer in the model, the process of noise injection can be described as follows. During each forward pass in the training phase, we first sample elements of noise matrix $\mathbf{\Omega}$ from standard normal distribution $\mathcal{N}(0, 1)$. Subsequently, the matrix $\mathbf{\Omega}$ is scaled with the quantization step size $\mathbf{\Delta}$. 
Finally, we add scaled noise to weights of non-salient columns  $\mathbf{W}_{[:, \textit{non-salient}]}$. The operation of the noise injection $\mho$ is given as 

\vspace{-0.5em}
\begin{align}
\label{eq:gni_finetuning}
\mho(\mathbf{W})=     \begin{cases}
       \mathbf{W_{[:,\textit{salient}]}}, \hspace{-.5em} \\
       \mathbf{W_{[:,\textit{non-salient}]}} + \frac{1}{2} \mathrm{diag}(\mathbf{\Delta}) \mathbf{\Omega} \hspace{-.5em} \\
     \end{cases},
\end{align}
where $\mathrm{diag}(\mathbf{\Delta})$ is the diagonal matrix with elements of the vector $\mathbf{\Delta}$.

Only weights of the salient columns $\mathbf{W_{[:,\textit{salient}]}}$ are updated during training, whereas weights of other columns $\mathbf{W_{[:,\textit{non-salient}]}}$ are frozen. We do not inject noise to salient weights since small perturbations in them can cause high model degradation. 

The quantization step size $\mathbf{\Delta}$ is determined only for weights in non-salient
columns $\mathbf{W}_{[:, \textit{non-salient}]}$. To closer match the initial distribution of the weights, quantization scale factors including in $\mathbf{\Delta}$ are estimated for each row individually. For $i$-s row the scale factor  $\Delta_{i}$ is computed as: 

\vspace{-0.5em}
\begin{equation}
    \label{eq:gni_finetuning_quant_step}
    \Delta_{i} = \frac{\alpha_{i}}{2^{b-1} - 1},
\end{equation}
where $b$ is the bit-width and $\alpha_{i}$ is the quantization parameter. As in quantization methods, smaller bit-width $b$ corresponds to higher quantization noise. The parameter $\alpha_i$ is estimated by optimizing weight error through linear search as discussed in Appendix~\ref{appendix:uniform quantization}.

Based on Equations~\ref{eq:gni_finetuning} and \ref{eq:gni_finetuning_quant_step}, the variance of the injected noise is determined by the distribution of non-salient weights across rows. We exclude salient columns from this distribution, as the salient weights may induce large quantization error and distort row-wise scale factors. This approach helps us to minimize the noise variance, which, in turn, leads to a reduction in the deviation of the non-salient weights during training. 

By sampling noise in such way we can use it for quantization pre-training experiments discussed in Section~\ref{sec:noise-pretrain}.
 


\begin{table*}[h]
\centering
\begin{tabular}{l|cc|cc|cc}
\toprule
& \multicolumn{2}{c|}{LLaMA2-7b} & \multicolumn{2}{c|}{LLaMA2-13b} & \multicolumn{2}{c}{LLaMA3-8b} \\
& T\"ULU-V2-mix       & OpenOrca       & T\"ULU-V2-mix       & OpenOrca       & T\"ULU-V2-mix       & OpenOrca       \\ 
\midrule
FT          & $71.97$    & \underline{$71.88$}    & \underline{$75.09$}    & \underline{$75.21$}    & \underline{$76.13$}    & $\mathbf{77.02}$    \\
LoRA        & $71.78$    & $70.89$    & $74.03$    & $74.01$    & $75.91$    & $75.63$    \\
DoRA        & \underline{$72.03$}    & $70.97$    & $73.97$    & $73.96$    & $75.89$    & $75.72$    \\
\midrule
\naming{} & $\mathbf{73.33}$    & $\mathbf{72.33}$    & $\mathbf{75.93}$    & $\mathbf{76.02}$    & $\mathbf{76.37}$    & \underline{$76.78$}    \\
\bottomrule
\end{tabular}
\caption{Mean accuracy of LLaMA models fine-tuned with various instructive datasets and different methods. }
\label{tab:fp_mean}
\end{table*}

\begin{table*}[h]
\centering
\begin{tabular}{llccc}
\hline
\textbf{Bits} & \textbf{Method} & \textbf{LLaMA2-7b} & \textbf{LLaMA2-13b} & \textbf{LLaMA3-8b} \\
\hline
\multirow{3}{*}{4 bit} 
& STE            & \underline{$72.43$} & $\mathbf{75.29}$ & \underline{$74.84$} \\
& QUIK + LORA    & $63.99$ & $71.08$ & $74.27$ \\
& \naming{}  & $\mathbf{72.53}$ & \underline{$74.50$} & $\mathbf{75.46}$ \\
\hline
\multirow{3}{*}{3 bit}
& STE            & \underline{$69.82$} & $\mathbf{74.37}$ & $70.24$ \\
& QUIK + LORA    & $62.91$ & $71.30$ & \underline{$71.65$} \\
& \naming{}  & $\mathbf{71.00}$    & \underline{$74.34$}    & $\mathbf{73.27}$ \\
\hline
\multirow{3}{*}{2 bit}
& STE            & \underline{$58.20$} & \underline{$62.19$} & $48.96$ \\
& QUIK + LORA    & $41.44$ & $47.14$ & \underline{$53.80$} \\
& \naming{}  & $\mathbf{61.09}$ & $\mathbf{67.61}$ & $\mathbf{58.89}$ \\
\hline
\end{tabular}
\caption{Mean accuracy of quantized and then fine-tuned models. For fine-tuning we used T\"ULU-V2-mix. }
\label{tab:quant_mean}
\end{table*}

\section{Experiments}
\label{sec:exps}
In this section, we describe the experimental procedure used to test the performance of \naming{} compared to others.

\subsection{Data}
\label{sec:exps_data}

Following previous studies \cite{nikdan2024rosa,hu2021lora,liu2024dora}, we focus on the instruction tuning task. For this purpose, we use the TULU-V2-Mix as the main source of data \cite{ivison2023camels}, as it encompasses a wide range of instructions from different sources. This dataset has been filtered, contains a substantial amount of data without being too large, and models tuned to this set show superior performance. Additionally, we utilize the OpenOrca dataset \cite{mukherjee2023orca} to demonstrate that our method does not depend on a specific set of instructions.

The sensitivity metrics to find salient columns are estimated based on 512 random sentences from the Pile validation dataset \cite{xiao2023smoothquant}.

\subsection{Baselines}

We consider several baselines for both full precision and quantized experiments. All baselines are applied to LLaMA2-7b, LLaMA2-13b and LLaMA3-8b.

\textbf{Full precision} version includes the choice of baselines, following recent studies \cite{liu2024dora,nikdan2024rosa}. We employ:
\begin{itemize}
    \item LoRA is a widely used adapter-based method \cite{hu2021lora}
    \item DoRA is modification of LoRA outperforming all current PEFT methods \cite{liu2024dora}
    \item FT is full fine-tuning of all parameters
\end{itemize}
We do not include PEFT methods connected with prompt tuning, as they show worse performance compared to adapter-based methods \cite{xu2023parameter}.

\textbf{Quantized} version is presented by baselines of only weight quantization at $\{4, 3, 2\}$ bit-widths:

\begin{itemize}
    \item STE is quantization-aware fine-tuning of all parameters of a pre-trained model \cite{bengio2013estimating}. During fine-tuning all parameters are trained, but 128 salient columns are updated in full-precision without quantization.
    \item QUIK + LoRA is an application of LoRA to the QUIK quantized model. Only low-rank adapters are trained, while the quantized weights and the salient weights are frozen. 
\end{itemize}

QUIK is a mixed-precision quantization method, that leverages GPTQ  for quantization non-salient columns, while keeping the salient weight in full-precision \cite{frantar2022gptq, ashkboos2023quik}. 
Due to the techniques, QUIK achieves the highest performance among PTQ methods, such as GTPQ \cite{frantar2022gptq}, AWQ \cite{lin2023awq}, SmoothQuant \cite{xiao2023smoothquant}.

\subsection{Evaluation and Datasets}

We perform a comprehensive evaluation measuring zero-shot performance on HellaSwag \cite{zellers2019hellaswag}, BoolQ \cite{clark2019boolq}, WinoGrande \cite{sakaguchi2021winogrande}, PiQA \cite{tata2003piqa}, ARC-easy, and ARC-challenge \cite{clark2018think} using the LM Eval Harness \cite{eval-harness}. The choice of baselines is similar to those in previous studies \cite{egiazarian2024extreme,frantar2022gptq,van2024gptvq}. 

We demonstrate average accuracy across all the datasets, detailed per-dataset comparison can be found in Section~\ref{appendix:bench_results}.

\subsection{Compute Budget}
In all experiments, the number of salient columns in the models is fixed at $128$. Furthermore, we fix our training budget at $500$ training iterations, unless specified otherwise. According to a recent study \cite{komatsuzaki2019one}, it is more effective to train for one epoch with a larger dataset rather than multiple epochs with less data. Therefore, all $500$ iterations are performed within one epoch with no instruction repetitions.

\subsection{Training Details}

The training was performed with 4 GPUs  ( 40 GB each) for 500 iterations. The batch size is 128 for 7b models and 64 for 13b models. For baseline methods, the learning rate was set to $3 \times 10^{-5}$ for LLaMA2 models and to $1 \times 10^{-5}$ for the LLaMA3 model. We experimented with different learning rates and found these to be the most beneficial for baseline methods. We used a cosine annealing scheduler with the warmup ratio of 0.03. The LoRA and DoRA alpha and dropout values were as specified in the original papers, and the rank was set to 64 to match the number of trainable parameters in our method. Thus, the number of trainable parameters is 160M for LLaMA2-7b, 250M for LLaMA2-13b, 167M for LLaMA3-8b.

For our method, the learning rate was set to $1 \times 10^{-4}$ for salient columns of LLaMA2 models and to $1 \times 10^{-5}$ of the LLaMA3 model. We fixed the number of salient columns at 128, such that the number of trainable parameters is 174M for LLaMA2-7b, 272M for LLaMA2-13b, and 176M for LLaMA3-8b.

In the case of full fune-tuning with the noise injection, the learning rate was set to $3 \times 10^{-5}$ and $1 \times 10^{-5}$ for LLaMA2 \& 3 models, correspondingly.

\begin{table}[h]
\centering
\resizebox{0.48\textwidth}{!}{
\begin{tabular}{lcccc}
\toprule
& \multicolumn{2}{c}{LLaMA2-7b} & \multicolumn{2}{c}{LLaMA2-13b} \\
& Performance & Compute$^\dagger$ & Performance & Compute$^\dagger$ \\ 
\midrule
T\"ULU2 & $73.49$ & $6.7$B / $5$K & $75.51$ & $13$B / $5$K \\ 
T\"ULU2-DPO & $73.8$ & $6.7$B / $5$K & $75.53$ & $13$B / $11$K \\ 
\midrule
\naming{} & $73.33$ & $174$M / $500$ & $75.93$ & $272$M / $500$ \\
\bottomrule
\end{tabular}
}
\caption{Comparison of Performance and Compute for LLaMA2 Models using our fine-tuning method versus original T\"ULU2 models. Note: Compute values are represented as Trainable Parameters / Iterations.}
\label{tab:tulu_compare}
\end{table}

\section{Results}
In this section, we present the results of our computational experiments and answer the questions posed in Section~\ref{sec:questions}. In short, our results are as follows:

\begin{itemize}
\item[\textbf{Q1:}] The results confirm that fine-tuning a subset of salient weights produces results comparable to those obtained using low-rank adapters.
\item[\textbf{Q2:}] Noise injections lead to improved model performance.
\item[\textbf{Q3:}] We could not confirm that models trained with noise injections are more robust to further degradation.
\end{itemize}

\subsection{Full Precision}
The average performance across evaluation benchmarks for full precision models is presented in Table~\ref{tab:fp_mean}. \naming{} generally shows superior metrics across most models and instruction sets. However, we observe slight underperformance in LLaMA3 on the OpenOrca subset, where full training proves superior. This issue likely stems from the choice of learning rate and schedule, which can impact the tuning of outliers.

\subsection{Quantized Models}
We present the averaged performance of models quantized with different precision (4, 3, 2) in Table~\ref{tab:quant_mean}. For 4 and 3 bits \naming{} achieves comparable quality with STE, however, latter one requires significantly more compute. In the 2-bit setting, \naming{} shows a substantial quality improvement, surpassing the second-ranked model by over 5 points. 

\subsection{Comparison with T\"ULU2}

We compare \naming{} with T\"ULU2 models \cite{ivison2023camels}, which are LLaMA2 models fine-tuned using a combination of instructions and DPO \cite{rafailov2023direct}. These models are among the top-performing LLaMA2 modifications but demand significant computational resources.

In Table~\ref{tab:tulu_compare}, we show that by applying \naming{} with significantly lower computational budget (a smaller number of parameters and iterations) we achieve comparable results for LLaMA2-7b and outperform T\"ULU2 for 13b.

\subsection{Scaling Properties}

We perform experiments to explore the performance of \naming{} and baselines with scaling data using LLaMA2 and LLaMA3 models. The results reported in Figure~\ref{fig:performance_comparison} show that while LoRA and DoRA exhibit unstable performance with scaling data, our method and full fine-tuning are more stable. Moreover, our method consistently ranks first across nearly all data budgets.

\begin{table}[h]
\centering
\begin{tabular}{lccc}
\hline
\textbf{Method} & \textbf{4 bit} & \textbf{3 bit} & \textbf{2 bit} \\
\hline
Salient FT             & $72.82$ & $\mathbf{71.06}$ & $59.82$ \\
Pre-\naming{}    & $\mathbf{73.15}$ &  $70.24$   & $47.08$ \\
Post-\naming{}   & $72.53$ & $71.00$    & $\mathbf{61.09}$ \\
\hline
\end{tabular}
\caption{Mean performance for quantized models with or without applying \naming{} before or after quantization, results are demonstrated for LLaMA2-7b model.}
\label{tab:pre_post_ablation}
\end{table}

\begin{table*}[ht]
    \centering
    \begin{tabular}{llccccc}
        \hline
        Bits & Model & $\Vert \mathbf{D}_{j} \Vert_{2}^{2} \Vert \mathbf{X}_{j} \Vert_{2}^{2}$ & $\Vert \mathbf{D}_{j} \Vert_{\infty} \Vert \mathbf{X}_{j} \Vert_{\infty}$ & $\Vert\mathbf{X}_{j}\Vert_{\infty} $ & $\Vert \mathbf{D}_{j} \Vert_{\infty} \Vert \mathbf{X}_{j} \Vert_{\infty}^{1/2}$ & $\Vert \mathbf{D}_{j} \Vert_{\infty} \Vert \mathbf{X}_{j} \Vert_{\infty}^{2}$ \\
        \midrule
        \multirow{4}{*}{4 bit} 
         & LLaMA2-7b & $\mathbf{69.86}$ & $69.85$ & $69.68$ & $69.55$ & $69.52$ \\
         & T\"ULU2-7b & $72.94$ & $\mathbf{73.17}$ & $72.77$ & $72.22$ &  $72.78$ \\
         \cline{2-7}
         & LLaMA2-13b & $72.92$ & $\mathbf{72.99}$ & $72.83$ & $72.83$ & $72.56$ \\
         & T\"ULU2-13b & $75.12$ & $74.86$ & $ 75.19$ & $\mathbf{75.47}$ & $75.17$ \\
        \midrule
        \multirow{4}{*}{3 bit} 
         & LLaMA2-7b & $67.50$ & $\mathbf{68.31}$ & $67.47$ & $68.09$ & $67.86$ \\
          & T\"ULU2-7b & $70.91$ & $\mathbf{71.30}$ & $70.85$ & $71.14$ & $70.88$ \\
          \cline{2-7}
         & LLaMA2-13b & $71.92$ & $71.59$ & $\mathbf{72.10}$ & $71.77$ & $71.45$ \\
         & T\"ULU2-13b & $\mathbf{74.33}$ & $74.07$ & $74.07$ & $74.09$ & $74.31$ \\
        \midrule
        \multirow{4}{*}{2 bit} 
         & LLaMA2-7b & $45.86$ & $46.78$ & $45.99$ & $46.81$ & $\mathbf{46.83}$ \\
         & T\"ULU2-7b & $\mathbf{54.84}$ & $46.85$ & $46.78$ & $48.56$ & $48.20$ \\
         \cline{2-7}
         & LLaMA2-13b & $57.07$ & $\mathbf{57.36}$ & $51.83$ & $57.30$ & $56.73$ \\
         & T\"ULU2-13b & $59.62$ & $59.62$ & $59.43$ & $\mathbf{60.67}$ & $59.39$ \\
        \hline
    \end{tabular}
    \caption{Performance of LLaMA2 and T\"ULU2 models after QUIK quantization with salient columns selected via various metrics. Weight perturbation is given by $\mathbf{D}_{j} = \mathbf{W}_{:, j} - Q(\mathbf{W}_{:, j})$.} 
    \label{tab:outlier_ablation}
\end{table*}
\begin{table}[h!]
\centering
\resizebox{0.48\textwidth}{!}{
    \begin{tabular}{lcccc}
    \toprule
    \multirow{2}{*}{Model} & \multicolumn{2}{c}{Outliers FT} & \multicolumn{2}{c}{Full FT} \\
    \cline{2-5}
     & w/ Noise & w/o Noise & w/ Noise & w/o Noise \\
    \midrule
    LLaMA2-7b & $\mathbf{73.33}$ & $73.16$ & $71.64$ & $71.97$ \\
    LLaMA2-13b & $\mathbf{75.93}$ & $74.80$ & $74.58$ & $75.09$ \\
    LLaMA3-8b & $\mathbf{76.37}$ & $75.45$ & $76.32$ & $76.13$ \\
    \bottomrule
    \end{tabular}
}
\caption{Mean Performance of LLaMA models with and without Noise Injection for outlier fine-tuning and full model fine-tuning}
\label{tab:noise_ablation}
\end{table}

\section{Ablation}

\subsection{Comparison sensitivity metrics}
We study sensitivity metrics with respect to different noise levels (various perturbation magnitudes), which translate into varying quantization precision. In this experiment, the non-salient weights of LLaMA2 and T\"ULU2 with 7B and 13B parameters.  Models are quantized with QUIK, the salient weights are not updated. We select 128 columns of salient weights.

Mean results for zero-shot tasks in Table~\ref{tab:outlier_ablation} show that for most precisions, the best performance is achieved with salient columns identified by Equation \ref{eq:our_metric} with $\gamma=1, \rho=\infty, \tau=\infty$ (second column). Columns identified by the squared $l_{2}$ norm of the input feature (the OWQ metric) show better performance only for T\"ULU2 quantized to 3 and 2 bits. 
Choosing salient columns solely by the input features (the QUIK metric) leads to underperformance, especially for 2 bit.
Therefore, identifying salient columns sensitive to quantization noise requires considering both the weight quantization error and the maximum values of input activation.


Based on the results, we chose the best-performing sensitivity metric with $\gamma=1, \rho=\infty, \tau=\infty$. However, the results do not reveal a clear rule for selecting the optimal sensitivity metric, as performance varies across different bit-widths and models with no discernible pattern. This remains an area for future research.

\subsection{Noise Injection Impact}
To ablate the importance of QNI in the full-precision setting, we measure the mean performance of LLaMA2 models with and without noise injections for both salient columns fine-tuning and full fine-tuning. In the latter case, the noise is applied to the entire weight matrix.

The results in Table~\ref{tab:noise_ablation} show that QNI consistently enhances the performance of outlier fine-tuning. Although QNI can reduce performance when applied to the entire network, it still benefits LLaMA3-8b. Notably, outlier fine-tuning outperforms full fine-tuning, but only when QNI is used.

\subsection{Quantization Before and After Training}
\label{sec:noise-pretrain}
From studies related to QAT, it is known that pre-training a model with noise injection enables to improve its predictive capabilities after quantization \cite{defossez2021differentiable, shvetsov2022quantnas}. 
Based on those observations, in this section we examine the performance of the quantized LLaMA2-7b after fine-tuning full precision salient columns
in several settings:
\begin{itemize}
    \item \textbf{Pre-\naming{}}. Applying \naming{} prior to the quantization.
    \item \textbf{Post-\naming{}}. Applying \naming{} after the quantization.
    \item \textbf{Salient FT}. Fine-tuning salient columns after quantization with no noise injected
\end{itemize} 

In the case of the pre-training, the bit-width for the model quantization corresponds to the noise level injected during the training. For the post-training, the noise injection is always performed at 4 bit.

Table~\ref{tab:pre_post_ablation} presents the average scores achieved by the models across evaluation benchmark. In the case of 4 bit quantization the Pre-\naming{} model considerable outperforms other models. But in the case of 3 and 2 bits,
fine-tuning salient columns after quantization enables to achieve quantized models better generative capabilities. 

It can be explained by significant deviation of the quantized weights from their original values that is induced by the extremely low-bit quantization. As a result, the interrelations between the salient weights and the quantized weights are disrupted, and the positive effect of pre-training disappears. However, post-training of the salient weight enables to form them new relations with other weights, so the model partially recovers its generative capabilities.

Also it can be observed that application of \textbf{Post-\naming{}} and \textbf{Salient FT} to model quantized in 3 bit gives the similar scores. But in the case of 2 bit quantization, the noise injection improves the fine-tuning of the quantized model.

\section{Conclusion}

In this paper, we introduce \naming{}, a parameter-efficient fine-tuning method that trains only weights in a small subset of salient columns while injecting quantization noise into the frozen weights. \naming{} proves to be superior to previous fine-tuning strategies in both full precision and quantized settings, requiring less compute budget. In data scaling experiments, \naming{} demonstrates greater stability than previous PEFT methods and outperforms both PEFT and full fine-tuning across nearly all data budgets. Our ablation studies show that QNI is beneficial but only with salient weights. Although \naming{} outperforms previous methods, further research is needed to determine how to maximize its performance in quantized settings.

We generalize the criterion for selecting salient columns from previous studies and empirically compare various parameters. Our experiments show that while some criteria perform better than others, none emerge as a clear dominant choice. This significant finding underscores the need for further research to refine these criteria.

\section{Limitations}
We find the main limitations of our work as follows:

\begin{enumerate}
    \item We report results of \naming{} exclusively for LLaMA models. Currently, numerous open-source pre-trained LLMs with high generative capabilities are available. However, LLaMA models are the most commonly chosen for studying the efficiency of modern PEFT and quantization methods. Despite the architectural similarities among most LLMs, future experiments with different models are necessary.
    \item For quantizing models, we use only the GPTQ method, which is widely used for mixed-precision quantization of LLMs. This method improves the performance of quantized models by aggregating quantization error into columns stored in full precision. However, \naming{} can be easily integrated with other methods, such as conventional RTN or QuantEase.
    \item Experiments with \naming{} report results for salient columns selected using the sensitivity metric (\ref{eq:our_metric}) with $\gamma=1$. Our proposed metric, based on our analysis, shows high sensitivity of the salient columns to quantization in most LLaMA2 cases. However, other sensitivity metrics may yield better performance for \naming{} and mixed-precision quantization in different LLMs.

    \item Noise parameters for fine-tuning the salient weights are determined using the QNI approach. However, other noise distributions may also enhance the fine-tuning process. 
    Identifying the optimal noise distribution is beyond the scope of this paper.

    \item In this study, we focus on developing the \naming{} algorithm for effective fine-tuning of LLMs, but we do not provide computationally efficient implementations of CUDA kernels for the algorithm. In the future, CUDA kernels for \naming{} can be developed based on the code from QUIK \cite{ashkboos2023quik} and OWQ \cite{lee2024owq}.

    \item We train \naming{} with only a few fine-tuning instruction sets, selected for their size and high benchmark results in previous studies. However, expanding the number of fine-tuning sets could make the experiments more comprehensive.

    \item We evaluate our method using six distinct benchmarks inherited from various previous studies. In future research, it would be beneficial to include more benchmarks to gain additional insights.

\end{enumerate}

\section{Potential Risks}
The \naming{} method poses risks similar to those of any PEFT method. 
For example, it omits explicit safety training measures, so could be applied to fine-tune LLMs for generating harmful content. Also it can be applied to tailor LLMs to tailor highly specific and potentially dangerous outputs.

\section{Acknowledgment}
The work was supported by the Analytical center under the RF Government (subsidy agreement 000000D730321P5Q0002, Grant No. 70-2021-00145 02.11.2021).

\bibliography{literature}

\clearpage
\appendix
\onecolumn

\section{Uniform quantization} \label{appendix:uniform quantization}
\begin{figure}
\centering
\begin{tikzpicture}
\begin{axis}[
    ylabel={$Q(w)$}, 
    xlabel={$\mathbf{w}$}, 
    xtick distance=0.25, 
    ytick distance=1, 
    grid=major, 
    ymin=0, ymax=4, 
    xmin=0, xmax=1, 
    axis lines=middle, 
    no markers, 
    samples=100, 
    domain=0:1, 
    ]

    \addplot+[const plot, thick, blue] coordinates {(0,1) (0.25,2) (0.5,3) (0.75,4) (1,4)};
\addplot+[mark=triangle, mark options={fill=green, draw=green}] coordinates {(0.75,3.2)} node[above, left, blue] { $\Delta$ step size};
\end{axis}
\end{tikzpicture}
\caption{Uniform quantization step function with real valued one dimensional $w$ and integer valued $Q(w)$.}
\label{fig:step_function}
\end{figure}
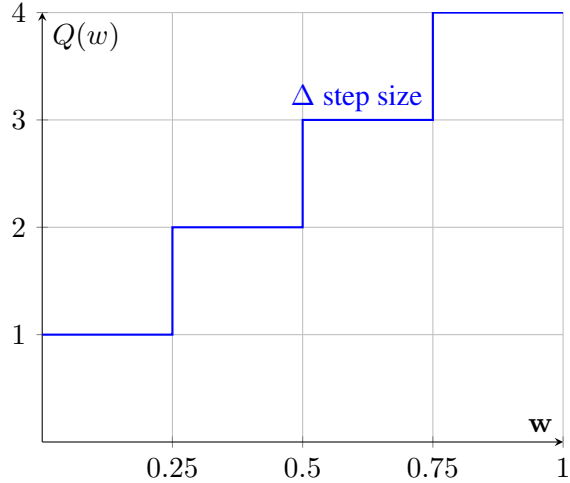

While non-uniform quantization may lead to higher compression rates, in our work we focus on uniform quantization since it widely used in efficient PTQ methods such as GPTQ, QUIK, OWQ \cite{frantar2022gptq, ashkboos2023quik, lee2024owq}. Quantization is a mapping  that converts a range of full-precision values into a discrete range of values allowing usage of integer arithmetic and reduced memory consumption. For example, Fig.~\ref{fig:step_function} depicts a mapping  with the quantization scale size $\Delta=\frac{1}{4}$ of float values from the interval $(0, 1)$ into integer values. 

In our work we apply  uniform symmetric quantization with the row-wise quantization step size $\mathbf{\Delta}$. In this case, 
computations of quantization, dequantization and estimation of $\mathbf{\Delta}$ are performed for the bit-width $b$ as below
\hspace{-0.5em}
\begin{gather}
\label{eq:quantization} 
q_{\text{min}}=-2^{b-1}, \quad q_{\text{max}}=2^{b-1}-1 \\
\text{clamp}(x; q_{\text{min}}, q_{\text{max}}) = \max(q_{\text{min}},\min(x,q_{\text{max}})) \\
\mathbf{\Delta} = (\Delta_{1}, \ldots, \Delta_{n})^{\mathrm{T}}, \quad  \Delta_{i} = \frac{\alpha_{i}}{q_{\text{max}}} \\
\mathbf{W}^{\text{int}}_{i, :}  = \text{clamp}\left(\left\lfloor\frac{\mathbf{W}_{i,:}}{\Delta_{i}} \right\rfloor; q_{\text{min}}, q_{\text{max}}\right) \\
  \mathbf{W} \approx Q(\mathbf{W}) = \mathrm{diag}(\mathbf{\Delta})\mathbf{W}^{\text{int}}
\end{gather}
where $\Delta_{i}$ is the scale factor for $i$ row $\mathbf{W}_{i,:}$, $\mathbf{W}^{int}$ denotes the matrix of the quantized weights, $\mathrm{diag}(\mathbf{\Delta})$ is the diagonal matrix with elements of the vector $\mathbf{\Delta}$. For the given bit-width $b$, the parameter $\alpha_{i}$ is found for each row by performing linear grid search over the interval $[0, \max(\mathbf{W}_{i, :})]$, where $\max(\mathbf{W}_{i, :})$ is the maximum element of $i$ row . The search is conducted to minimize layer-wise mean squared error between weights:
\hspace{-0.5em}
\begin{equation}
  \label{eq:layerwise_quant_error}
  \mathrm{argmin}_{\mathbf{\Delta}} \Vert \mathbf{W} - Q(\mathbf{W})\Vert_{2}^{2},
\end{equation}

\section{Details of LLMs quantization} \label{appendix:LLM quantization}
For only weight quantization of LLaMA and T\"ULU2 models models, we apply QUIK implementation of mixed-precision GPTQ method \cite{ashkboos2023quik, frantar2022gptq}. We isolate 128 salient columns in full-precision. Non-salient columns are subjected to uniform symmetric quantization, as discussed in Appendix \ref{appendix:uniform quantization}. The salient columns are identified through sensitive metrics described in Section \ref{sec:metrics}. The Hessian matrix for the GPTQ method is computed on 
128 random samples of the Wikitext-2 dataset.

\section{Straight Through Estimator} \label{appendix: STE}
STE can be described in two steps:

\begin{itemize}
    \item  Obtain quantized weights $Q(\mathbf{W})$ from the real-valued parameters $\mathbf{W}$ with some quantization function $Q$, which is usually is non differentiable.
    \item Compute gradients at quantized weights $Q(\mathbf{W})$ and update real valued weights  $\mathbf{W}_{t+1} \leftarrow \mathbf{W}_{t} - \tau \nabla f(Q(\mathbf{W}))$
\end{itemize}
STE makes a particular choice of a quantization function to obtain the discrete weights from the real-valued weights. This approximation can be justified in some settings \citep{lin2017towards} but in general the reasons behind its effectiveness are unknown.

\section{Detailed Benchmark Results} \label{appendix:bench_results}
In this section we report detailed benchmark results for LLaMA 2 \& 3 after training with different methods. Tables \ref{tab:appendix_tulu}, \ref{tab:appendix_orca} present accuracy metrics which are achieved by the full-precision models after fine-tuning on the T\"ULU-V2-mix and OpenOrca subsets. Corresponding mean values are listed in Table ~\ref{tab:fp_mean}. Tables present accuracy metrics which are achieved by quantized in {4, 3, 2} bits models after fine-tuning on the T\"ULU-V2-mix subset. Corresponding mean values are listed in Table ~\ref{tab:quant_mean}.

\begin{table*}[h]
\centering
\begin{tabular}{lccccccc}
\hline
\textbf{Model} & \textbf{Method} & \textbf{BoolQ} & \textbf{HellaSwag} & \textbf{WinoGrande} & \textbf{ARC-e} & \textbf{ARC-c} & \textbf{PiQA} \\
\hline
\multirow{4}{*}{LLaMA2-7b}
& FP        & $78.65$ & $\mathbf{76.91}$ & $69.93$ & $77.99$ & $48.63$ & $79.71$\\
& LoRA      & $80.28$ & $76.67$ & $69.85$ & $76.64$ & $47.95$ & $79.27$ \\
& DoRA      & $81.93$ & $76.27$ & $70.09$ & $76.05$ & $48.89$ & $78.94$ \\
& \naming{}   & $\mathbf{82.63}$ & $76.68$ & $\mathbf{70.80}$ & $\mathbf{80.01}$ & $\mathbf{49.91}$ & $\mathbf{79.92}$    \\
\hline
\multirow{4}{*}{LLaMA2-13b}
& FP        & $83.27$ & $79.77$ & $72.69$ & $80.43$ & $53.67$ & $\mathbf{80.69}$ \\
& LoRA      & $81.10$ & $79.57$ & $72.77$ & $78.91$ & $51.28$ & $80.52$ \\
& DoRA      & $81.01$ & $79.64$ & $72.22$ & $78.87$ & $51.54$ & $80.52$ \\
& \naming{}   & $\mathbf{84.22}$ & $\mathbf{80.18}$ & $\mathbf{73.24}$ & $\mathbf{82.20}$ & $\mathbf{55.38}$ & $80.36$ \\
\hline
\multirow{4}{*}{LLaMA3-8b}
& FP        & $83.64$ & $79.56$ & $74.35$ & $\mathbf{82.41}$ & $55.72$ & $81.12$  \\
& LoRA      & $83.30$ & $79.62$ & $75.14$ & $80.15$ & $56.06$ & $81.18$ \\
& DoRA      & $83.61$ & $79.53$ & $\mathbf{75.45}$ & $80.09$ & $55.63$ & $81.01$ \\
& \naming{}  & $\mathbf{83.88}$ & $\mathbf{80.02}$ & $75.22$ & $80.56$ & $\mathbf{57.00}$ & $\mathbf{81.56}$\\
\hline
\end{tabular}
\caption{LLaMA models performance fine-tuned with T\"ULU-V2-mix subset}
\label{tab:appendix_tulu}
\end{table*}

\begin{table*}[h]
\centering
\begin{tabular}{lccccccc}
\hline
\textbf{Model} & \textbf{Method} & \textbf{BoolQ} & \textbf{HellaSwag} & \textbf{WinoGrande} & \textbf{ARC-e} & \textbf{ARC-c} & \textbf{PiQA} \\
\hline
\multirow{4}{*}{LLaMA2-7b}
& FT        & $80.03$ & $\mathbf{77.02}$ & $69.69$ & $\mathbf{76.64}$ & $48.72$ & $79.16$ \\
& LoRA      & $78.81$ & $76.24$ & $68.82$ & $75.42$ & $46.59$ & $\mathbf{79.43}$ \\
& DoRA      & $78.78$ & $76.30$ & $68.92$ & $75.67$ & $46.93$ & $79.22$ \\
& Our Best  & $\mathbf{82.51}$ & $76.64$ & $\mathbf{72.22}$ & $74.71$ & $\mathbf{48.89}$ & $79.00$  \\
\hline
\multirow{4}{*}{LLaMA2-13b}
& FT        & $82.66$ & $\mathbf{80.30}$ & $73.01$ & $79.97$ & $54.78$ & $\mathbf{80.52}$ \\
& LoRA      & $81.68$ & $79.64$ & $72.85$ & $78.41$ & $51.11$ & $80.36$ \\
& DoRA      & $81.65$ & $79.64$ & $72.93$ & $78.28$ & $51.19$ & $80.09$ \\
& Our Best  & $\mathbf{85.44}$ & $80.07$ & $\mathbf{74.03}$ & $\mathbf{79.97}$ & $\mathbf{56.48}$ & $80.14$ \\
\hline
\multirow{4}{*}{LLaMA3-8b}
& FT        & $\mathbf{84.37}$ & $\mathbf{80.11}$ & $\mathbf{75.93}$ & $\mathbf{81.82}$ & $\mathbf{57.85}$ & $\mathbf{82.05}$  \\
& LoRA      & $82.84$ & $79.76$ & $74.19$ & $80.30$ & $55.54$ & $81.12$ \\
& DoRA      & $82.63$ & $79.71$ & $75.22$ & $80.30$ & $55.46$ & $81.01$ \\
& Our Best  & $84.34$ & $80.10$ & $75.53$ & $81.06$ & $57.76$ & $81.88$\\
\hline
\end{tabular}
\caption{LLaMA models performance fine-tuned with OpenOrca}
\label{tab:appendix_orca}
\end{table*}

\begin{table*}[ht]
    \centering
    \begin{tabular}{lllcccccc}
        \toprule
        \multirow{3}{*}{Bits} & \multirow{3}{*}{Model} & \multirow{3}{*}{Method} & \multicolumn{6}{c}{Benchmarks} \\
        \cline{4-9}
         & & & BoolQ & HellaSwag & WinoGrande & ARC-e & ARC-c & PiQA \\
        \midrule
        \multirow{9}{*}{4 bit} 
         & \multirow{3}{*}{LLaMA2-7b} & STE & $80.21$ & $\mathbf{76.27}$ & $70.01$ & $79.63$ & $48.55$ & $\mathbf{79.92}$ \\
         & & QUIK + LORA & $68.96$ & $74.85$ & $69.85$ & $55.30$ & $37.20$ & $77.80$ \\
         & & \naming{}  & $\mathbf{82.78}$ & $76.14$ & $\mathbf{70.48}$ & $\mathbf{79.76}$ & $\mathbf{50.00}$ & $79.71$ \\
        \cline{2-9}
         & \multirow{3}{*}{LLaMA2-13b} & STE & $\mathbf{84.77}$ & $79.16$ & $72.69$ & $\mathbf{80.76}$ & $\mathbf{53.67}$ & $\mathbf{80.69}$ \\
         & & QUIK + LORA & $74.89$ & $78.01$ & $72.22$ & $71.76$ & $50.17$ & $79.43$ \\
         & & \naming{}  & $84.65$ & $\mathbf{79.59}$ & $\mathbf{73.01}$ & $78.37$ & $53.50$ & $80.52$ \\
        \cline{2-9}
         & \multirow{3}{*}{LLaMA3-8b} & STE & $81.59$ & $78.55$ & $73.88$ & $79.76$ & $54.27$ & $81.01$ \\
         & & QUIK + LORA & $82.51$ & $77.73$ & $\mathbf{74.66}$ & $79.04$ & $51.62$ & $80.03$ \\
         & & \naming{}  & $\mathbf{83.15}$ & $\mathbf{79.05}$ & $74.09$ & $\mathbf{80.01}$ & $\mathbf{55.20}$ & $\mathbf{81.28}$ \\
        \midrule
        \multirow{9}{*}{3 bit} 
         & \multirow{3}{*}{LLaMA2-7b} & STE & $76.79$ & $74.19$ & $68.19$ & $75.04$ & $45.65$ & $79.05$ \\
         & & QUIK + LORA & $63.88$ & $72.00$ & $66.93$ & $61.24$ & $38.74$ & $74.64$ \\
         & & \naming{}  & $\mathbf{80.46}$ & $\mathbf{74.20}$ & $\mathbf{68.90}$ & $\mathbf{75.88}$ & $\mathbf{47.35}$ & $\mathbf{79.22}$ \\
        \cline{2-9}
         & \multirow{3}{*}{LLaMA2-13b} & STE & $83.33$ & $78.02$ & $71.59$ & $\mathbf{79.92}$ & $\mathbf{53.24}$ & $\mathbf{80.09}$ \\
         & & QUIK + LORA & $82.02$ & $76.64$ & $70.95$ & $71.51$ & $48.21$ & $78.45$ \\
         & & \naming{}  & $\mathbf{85.44}$ & $\mathbf{78.20}$ & $\mathbf{71.90}$ & $79.12$ & $51.54$ & $79.82$ \\
        \cline{2-9}
         & \multirow{3}{*}{LLaMA3-8b} & STE & $75.87$ & $74.38$ & $69.14$ & $74.41$ & $49.32$ & $78.29$ \\
         & & QUIK + LORA & $78.72$ & $74.54$ & $70.72$ & $77.31$ & $50.60$ & $78.02$ \\
         & & \naming{}  & $\mathbf{80.31}$ & $\mathbf{75.98}$ & $\mathbf{71.51}$ & $\mathbf{79.63}$ & $\mathbf{52.99}$ & $\mathbf{79.22}$ \\
        \midrule
        \multirow{9}{*}{2 bit} 
         & \multirow{3}{*}{LLaMA2-7b} & STE & $68.47$ & $58.90$ & $60.62$ & $57.66$ & $32.17$ & $71.38$ \\
         & & QUIK + LORA & $62.11$ & $26.77$ & $49.88$ & $29.67$ & $26.45$ & $53.75$ \\
         & & \naming{}  & $\mathbf{71.90}$ & $\mathbf{64.18}$ & $\mathbf{62.59}$ & $\mathbf{61.57}$ & $\mathbf{34.90}$ & $\mathbf{71.38}$ \\
        \cline{2-9}
         & \multirow{3}{*}{LLaMA2-13b} & STE & $73.09$ & $63.74$ & $61.40$ & $64.14$ & $36.09$ & $74.70$ \\
         & & QUIK + LORA & $59.36$ & $41.34$ & $55.41$ & $40.28$ & $27.82$ & $58.60$ \\
         & & \naming{} & $\mathbf{81.99}$ & $\mathbf{69.49}$ & $\mathbf{65.43}$ & $\mathbf{70.33}$ & $\mathbf{43.17}$ & $\mathbf{75.24}$ \\
        \cline{2-9}
         & \multirow{3}{*}{LLaMA3-8b} & STE & $60.46$ & $43.82$ & $54.46$ & $44.23$ & $27.65$ & $63.16$ \\
         & & QUIK + LORA & $64.68$ & $48.55$ & $58.25$ & $53.32$ & $32.17$ & $65.83$ \\
         & & \naming{} & $\mathbf{74.13}$ & $\mathbf{48.92}$ & $\mathbf{58.88}$ & $\mathbf{63.17}$ & $\mathbf{37.88}$ & $\mathbf{70.35}$ \\
        \bottomrule
    \end{tabular}
    \caption{Performance of quantized LLaMA models fine-tuned with T\"ULU-V2-mix subset}
    \label{tab:results}
\end{table*}

\clearpage

\section{T\"ULU-V2-mix subset}
\begin{figure*}
    \centering
    \includegraphics[width=\textwidth]{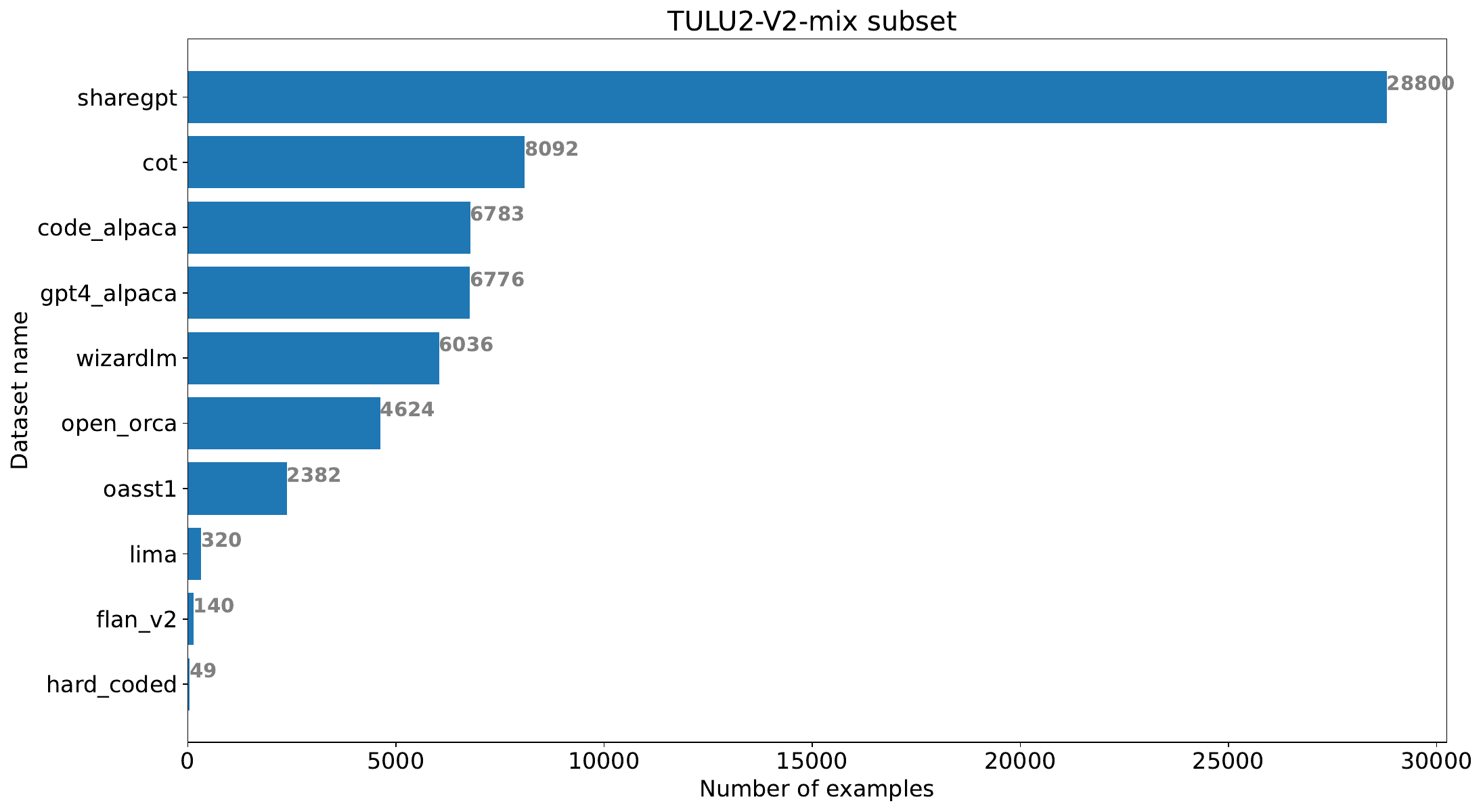}
    \caption{Number of examples in datasets included in T\"ULU-V2-mix subset}
    \label{fig:tulu_v2_mix_subset}
\end{figure*}

Figure \ref{fig:tulu_v2_mix_subset} shows number of examples in datasets included in the T\"ULU-V2-mix subset, which is used for fine-tuning experiments presented in this paper.

\end{document}